\documentclass[letterpaper, 10 pt, conference]{ieeeconf}  

\IEEEoverridecommandlockouts                              

\title{\LARGE \bf
Layer Decomposition and Morphological Reconstruction for Task-Oriented Infrared Image Enhancement

}

\author{Siyuan Chai, Xiaodong Guo and Tong Liu
\thanks{*This work was supported by National Natural Science Foundation of China under Grant 62476026. (Corresponding author: \textit{Tong Liu}).}
\thanks{Siyuan Chai, Xiaodong Guo and Tong Liu are with the School of Automation, Beijing Institute of Technology, Beijing 100081, China (e-mail: {\tt\small liutong2002@bit.edu.cn}, {\tt\small 3120230802@bit.edu.cn}).
        }%
}

\usepackage{booktabs}   
\usepackage{multirow}   
\usepackage{xcolor}     
\usepackage{graphicx} 
\graphicspath{{figs/}}    
\usepackage{amsmath} 
\usepackage[utf8]{inputenc}
\usepackage{amssymb} 
\usepackage{siunitx}     

\begin{document}

\maketitle
\thispagestyle{empty}
\pagestyle{empty}

\begin{abstract}

Infrared image helps improve the perception capabilities of autonomous driving in complex weather conditions such as fog, rain, and low light. However, infrared image often suffers from low contrast, especially in non-heat-emitting targets like bicycles, which significantly affects the performance of downstream high-level vision tasks. Furthermore, achieving contrast enhancement without amplifying noise and losing important information remains a challenge.
To address these challenges, we propose a task-oriented infrared image enhancement method. Our approach consists of two key components: layer decomposition and saliency information extraction. First, we design an $\boldsymbol{l_0}$-$\boldsymbol{l_1}$ layer decomposition method for infrared images, which enhances scene details while preserving dark region features, providing more features for subsequent saliency information extraction. Then, we propose a morphological reconstruction-based saliency extraction method that effectively extracts and enhances target information without amplifying noise.
Our method improves the image quality for object detection and semantic segmentation tasks. Extensive experiments demonstrate that our approach outperforms state-of-the-art methods.

\end{abstract}

\section{Introduction}

The environmental perception system in autonomous driving faces significant challenges under complex weather conditions, including fog, rain, snow, and low light.
Infrared imaging technology captures the radiation characteristics of targets and demonstrates strong resistance to interference under such challenging conditions.
Therefore, integrating infrared images with RGB can improve the perception of unmanned platforms under adverse weather conditions.

However, due to the limitations of the infrared imaging mechanism and the influence of external environments, infrared images may exhibit undesirable visual effects, such as low contrast. This makes the detection of targets, especially non-heat-emitting ones like bicycles, particularly challenging.
Methods based on Histogram Equalization (HE) \cite{stark2000adaptive, lu2021effective} modify the image histogram through a series of weighting or optimization methods to improve contrast. These methods focus on adjusting the global or local gray-level distribution, which makes it difficult to effectively perceive important information in the image and renders them insensitive to enhancing the contrast of infrared targets. 
   \begin{figure}[h]
    \centering
    \vspace{0.3cm}
        \includegraphics[width=1.0\columnwidth]{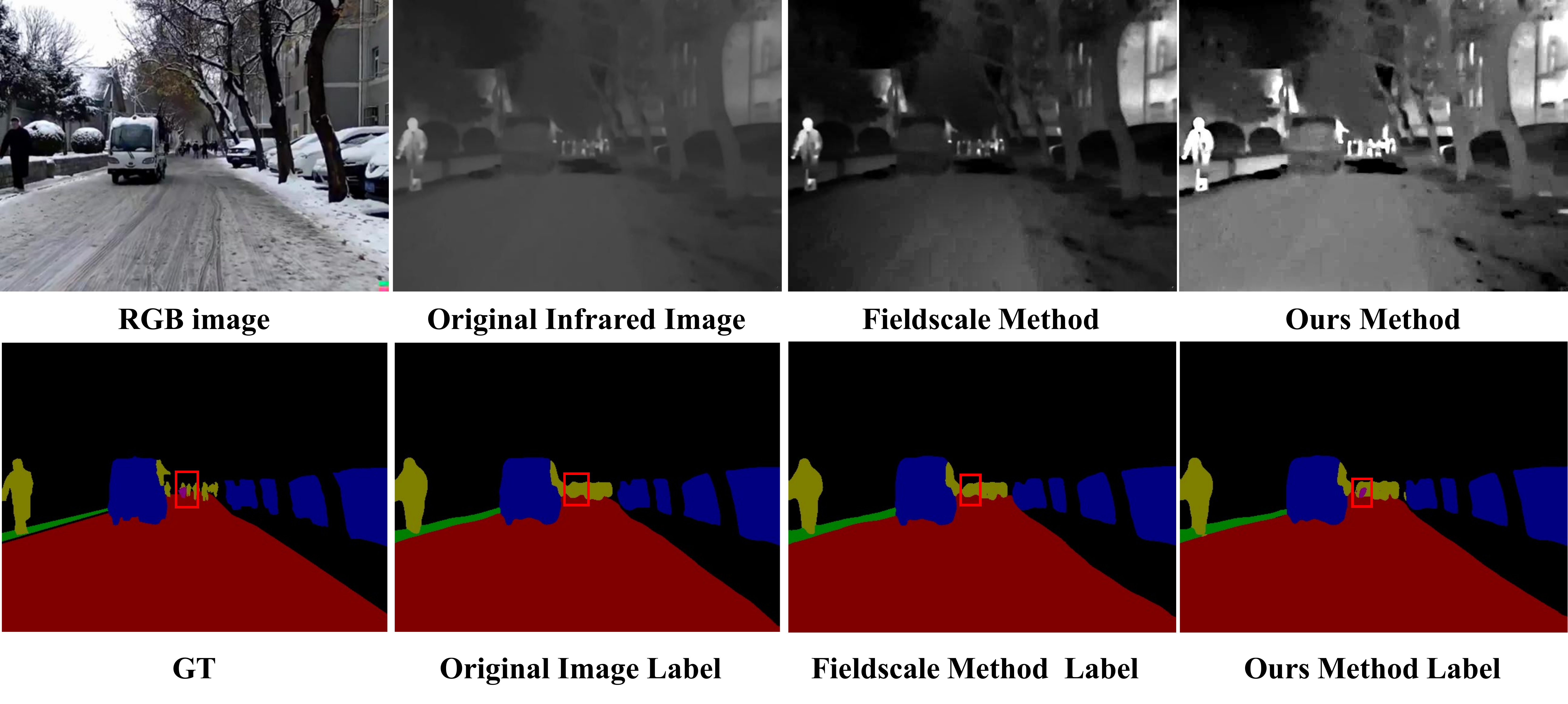}
    \caption{Visualization results of our infrared image enhancement algorithm and the Fieldscale method on the SUS semantic segmentation dataset. The areas highlighted by the red boxes demonstrate the superiority of our proposed method in high-level visual tasks.}
    \label{figurelabel}
    \vspace{-0.3cm}
\end{figure}
Methods based on the Human Visual System (HVS) \cite{zhao2014fast, katirciouglu2019infrared, li2021infrared} utilize the selective attention mechanism of the human visual system, applying different decomposition methods to enhance the important regions of the image. These methods typically add prominent bright areas to the original image, not only overlooking low-contrast targets in the dark regions but also potentially amplifying noise components unintentionally. 
Deep learning-based methods, evolving from Convolutional Neural Networks (CNNs) \cite{choi2016thermal} to Generative Adversarial Networks (GANs) \cite{marnissi2023gan}, generally produce impressive visual effects. However, their performance heavily relies on large training datasets and may suffers from over-enhancement, leading to a significant loss of features in the dark regions and complicating the detection of low-contrast targets.
Although existing methods can improve visual perception, they still face limitations in suppressing noise, preserving features in dark regions, and effectively highlighting scene targets. This can lead to errors in target recognition and classification in subsequent high-level vision tasks.
Therefore, the challenge lies in performing contrast enhancement without amplifying noise or losing important information.

\begin{figure*}[thpb]
    \centering
    \vspace{0.3cm}
    \includegraphics[width=0.8\textwidth]{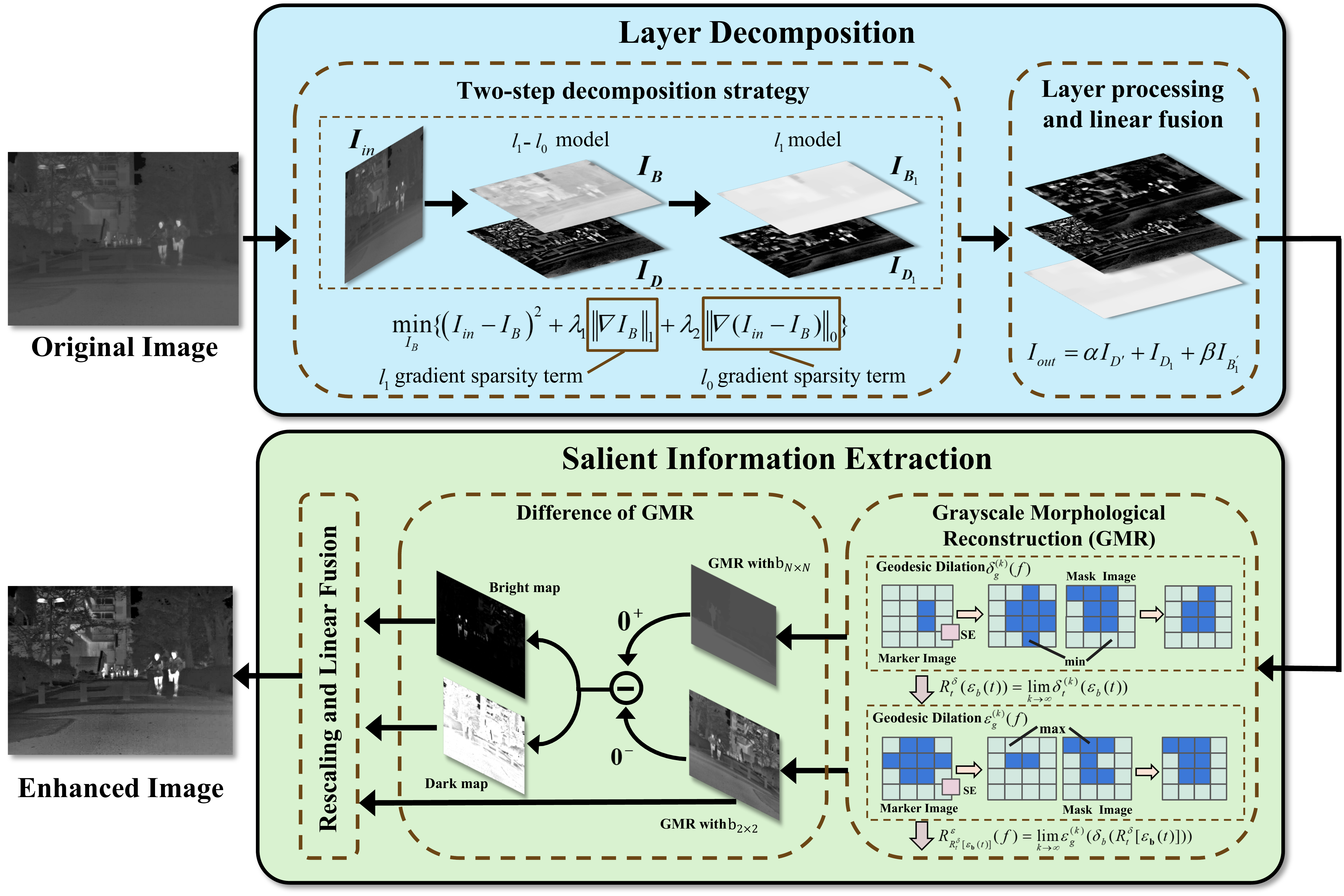}
    \caption{An overview of the framework for the proposed infrared image enhancement method is presented here. The framework comprises two key components: layer decomposition and salient information extraction. Layer decomposition employs a two-step decomposition strategy to enrich scene details while preserving important information. Subsequently, salient information extraction utilizes the differential results from GMR with two different structuring elements to extract and enhance targets in the scene.}
    \label{figurelabel}
\end{figure*}
In addition, enhanced infrared images should not only highlight prominent targets and rich details but also be suitable for advanced vision tasks. However, current infrared image enhancement algorithms primarily focus on improving visual quality and statistical metrics, often neglecting the requirements of high-level vision tasks, such as object detection or semantic segmentation. Meanwhile, existing task-oriented methods \cite{marnissi2021thermal} typically focus on enhancing high-contrast objects, such as pedestrians, in infrared images, often overlooking other targets present in the scene.

To address the aforementioned issues, we propose a task-oriented infrared image enhancement method. Here, "task-oriented" means designing the method based on the requirements of high-level vision tasks, such as object detection and semantic segmentation, with a focus on enhancing targets including pedestrians, vehicles, and bicycles. First, to reduce noise introduced by layer decomposition and suppress edge artifacts, we design an $l_0$-$l_1$ layer decomposition method for infrared images. This method introduces an $l_0$ gradient sparsity term in the detail layer and an $l_1$ gradient sparsity term in the base layer, enhancing scene details while preserving important information, such as features in dark regions. This enables subsequent saliency information extraction methods to capture more features.  
Next, leveraging the ability of morphological reconstruction techniques to remove uniformly connected regions with gray features, preserve edge contours and shapes, and suppress noise, we propose a morphological reconstruction-based method for extracting target saliency information. By applying differential grayscale morphological reconstruction operations, we effectively extract and enhance target information from the scene without amplifying noise. The resulting infrared image is more suitable for high-level visual tasks, as shown in Fig. 1.

Our main contributions include:

\begin{itemize}

\item We design an $l_0$-$l_1$ layer decomposition method for infrared images that enhances scene details, preserves important information, and provides more features for subsequent saliency information extraction methods.
\item We propose a saliency information extraction method based on morphological reconstruction, which effectively extracts and enhances target information from the scene without amplifying noise, thereby facilitating perception for advanced vision tasks.
\item We introduce a task-oriented infrared enhancement algorithm and evaluate its effectiveness based on the precision of high-level vision tasks. Extensive experiments on two standard datasets for object detection and semantic segmentation demonstrate that our method outperforms state-of-the-art (SOTA) methods.

\end{itemize}

\section{Related Works}

\subsection{Infrared Image Enhancement} 
Existing infrared image enhancement algorithms can be broadly categorized into two groups: Deep Learning Methods and Model-driven Methods.

\textit{1) Deep Learning Methods  }:
Recent advancements in deep learning, particularly methods based on CNN and GAN, have significantly improved infrared image enhancement \cite{fan2018dim, zhu2024mmff}. Choi et al. \cite{choi2016thermal} introduced a CNN-based thermal image enhancement method, and Kuang et al. \cite{kuang2019single} improved it with IE-CGAN to reduce background noise. Marnissi et al. proposed the TE-GAN method for contrast enhancement and denoising, later integrating the Vision Transformer (ViT) into TE-VGAN. Although deep learning methods are highly effective, they require large datasets and high computational power. Their training results are not always stable, and they often suffer from over-enhancement, leading to severe loss of features in dark areas and limiting their real-time applicability.

\textit{2) Model-driven Methods  }:
Model-driven methods require fewer resources and are more suitable for real-time applications. These methods primarily include techniques based on histogram equalization (HE) and the human visual system (HVS).  
HE-based methods \cite{mohan2012modified} modify the image histogram through weighting or optimization techniques to expand the dynamic range of low-contrast images. Weighted HE methods \cite{paul2020adaptive, wang2021range} improve classic HE by introducing local adaptation but often amplify noise and overexposure. Optimization-based HE methods \cite{liu2019optimized, zhang2022infrared} address this issue by focusing on local contrast but still require significant computational time.  
HVS-based methods \cite{li2025contrast} enhance infrared images by focusing on regions of interest through selective attention mechanisms. Chen et al. \cite{chen2021saliency} proposed a multi-scale saliency fusion method incorporating Gaussian smoothing and saliency detection. Wang et al. \cite{wang2024raw} applied inverted gray level mapping (IGIM) and gamma correction, and Gil et al. \cite{gil2024fieldscale} used grid min/max pooling to handle large temperature differences. These methods can effectively highlight salient objects in the scene but often overlook low-contrast targets in dark areas or over-enhance certain targets, thereby limiting their effectiveness. Therefore, we propose a target-oriented infrared image enhancement method that achieves contrast enhancement without amplifying noise or losing important information.

\subsection{Layer Decomposition } 

Layer decomposition methods \cite{zuo2011display, gu2012local, chen2020real} enhance image details by decomposing the image into a base layer and a detail layer. The process involves layer decomposition, base layer compression, detail layer enhancement, and finally fusing the base layer with one or more detail layers to obtain the enhanced image. Common tools for layer decomposition include Gaussian filters, bilateral filters, guided filters, and norm-based constraints. Although these methods can effectively improve image enhancement quality, they fail to simultaneously achieve effective noise suppression and edge enhancement. Therefore, we design an $l_0$-$l_1$ layer decomposition method specifically tailored for infrared images to reduce noise caused by layer decomposition and suppress edge artifacts.

\subsection{Morphological Reconstruction} 

Morphological reconstruction helps solve many challenging problems in image processing, including image dehazing \cite{salazar2018fast, ganguly2021single} and noise suppression \cite{wang2020residual}. This is attributed to the excellent characteristics of morphological reconstruction, which can effectively remove uniform connected regions with gray features while preserving good edge contours and shapes. Additionally, infrared image enhancement methods using mathematical morphology to enhance the contrast of prominent regions have been developed, including multi-scale top-hat transformations \cite{bai2012image} and multi-scale sequential switching operators \cite{bai2015morphological}. Therefore, we incorporate morphological reconstruction into our infrared image enhancement method and propose a morphological reconstruction-based target saliency information extraction method, which effectively extracts and enhances target information in the scene without amplifying noise.

\begin{figure*}[h!]
    \centering
    \vspace{0.3cm}
    \includegraphics[width=0.9\textwidth]{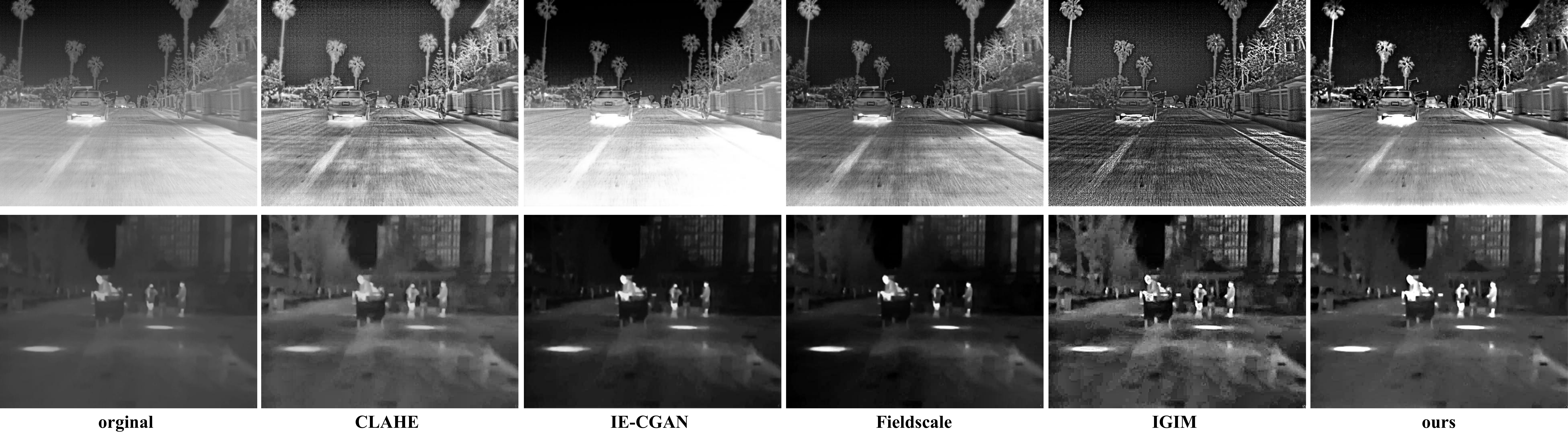}
    \caption{Qualitative comparison results of various methods on FLIR dataset and SUS dataset.The image above is from the FLIR dataset, while the image below is from the SUS dataset.}
    \label{figurelabel}
\end{figure*}

\section{Method}

\subsection{Overview Architecture}
In this section, we present the details of our proposed algorithm. Our infrared image enhancement framework consists of two key components: Layer Decomposition and Salient Information Extraction, focusing on detail enhancement and target information extraction, respectively. We adopt a two-step decomposition strategy, decomposing the infrared image into a compressed base layer and two enhanced detail layers. This approach results in an infrared image with enhanced target edges and rich scene details, thereby providing more features for subsequent saliency information extraction methods. Next, grayscale morphological reconstruction (GMR) techniques are used to eliminate uniform bright/dark connected regions in the infrared image. A differential algorithm is then applied to generate bright/dark saliency maps. Finally, the extracted target saliency information is fused with the base layer of the original image, resulting in an infrared image with highlighted targets. Each component is optimized to improve the visual quality of infrared images and to enhance the performance of high-level vision tasks. An overview of the framework for our proposed IR image enhancement method is illustrated in Fig. 2.

\subsection{Layer Decomposition}

To emphasize the targets within the image, enrich scene details, and preserve critical information such as features in darker regions, it is essential to enhance the subsequent extraction of salient information. This process is crucial for achieving improved overall results. Inspired by Liang et al. \cite{liang2018hybrid}, we propose an $l_0$-$l_1$ layer decomposition method tailored specifically for infrared images. The method employs an $l_0$ gradient sparsity term on the detail layer to establish structural priors, effectively preserving critical structural information. Simultaneously, an $l_1$ gradient sparsity term is applied to the base layer to maintain edge sharpness while suppressing unwanted artifacts. We implement a two-step decomposition strategy to decompose the infrared image into one base layer and two detail layers. The final output is an infrared image with enhanced target edges and enriched scene details, striking a balance between structural preservation and artifact suppression.

Our $l_0$-$l_1$ layer decomposition method formulates the image decomposition as an optimization problem, with the cost function defined as follows:
\begin{equation}\min_{I_B}\{\left(I_{in}-I_B\right)^2+\lambda_1\left\|\nabla I_B\right\|_1+\lambda_2\left\|\nabla(I_{in}-I_B)\right\|_0\},\end{equation}
where $I_{in}$, $I_B$, and $I_{in}-I_B$ denote the original image, base layer, and detail layer, respectively. \( \lambda_1 \) and \( \lambda_2 \) are constraint factors. The $l_1$ sparsity term is applied to the base layer to preserve larger gradients, whereas the $l_0$ sparsity term is applied to the detail layer. Due to the strict piecewise constant properties of $l_0$, large structures (edges) in the detail layer remain intact, whereas small textures (noise) tend to zero. As a result, the detail layer becomes more uniform, and edges are free from artifacts.

The solution to this model is obtained using the Alternating Direction Method of Multipliers (ADMM), yielding the base layer $I_B$. The detail layer $I_D$ is subsequently obtained by subtracting the base layer from the input image. As a result, structural information is retained in the detail layer $I_D$, whereas the primary texture information is transferred to the base layer $I_B$.

To further enhance the image, we adopt a dual-scale decomposition strategy. In the second scale decomposition, aimed at extracting richer texture and detail information, we use the $l_1$ model to perform secondary decomposition on the base layer obtained from the first decomposition. The base layer generated by the $l_0$-$l_1$ model undergoes iterative refinement, splitting the image into a compressed base layer and two enhanced detail layers. Specifically, we compress the base layer to suppress low-frequency redundancy while amplifying high-frequency components in the detail layers. The final enhanced image is expressed as:
\begin{equation}I_{out}=\alpha I_{D^{\prime}}+I_{D_1}+\beta I_{B_1^{\prime}},\end{equation}
where $I_{out}$ is the fusion result, $I_{D^{\prime}}$ and $I_{B_1^{\prime}}$ are the compressed and stretched forms of $I_{D}$ and $I_{B_1}$, and \( \alpha \) and \( \beta \) are the fusion coefficients for the base and detail layers. These coefficients are adjustable and selected based on experience.

\subsection{Salient Information Extraction}
To enhance the prominence of pedestrians, vehicles, and bicycles, which are critical for downstream high-level vision tasks, we propose a grayscale morphological reconstruction-based salient target extraction method.The Grayscale Morphological Reconstruction (GMR) technique effectively eliminates bright-dark connected regions in the infrared image. Subsequently, we employ a differential algorithm to generate bright/dark salient maps, extracting key information about targets required for high-level vision tasks without amplifying noise. Finally, the extracted salient target information is fused with the base layer of the original image to generate an infrared image with enhanced targets, which is more suitable for high-level vision tasks.

\textit{1) Fundamental Principles of Morphological Reconstruction}:
Mathematical morphology consists of two fundamental components: structuring elements (SE) and basic morphological operations. Each morphological operation is performed at the intersection of the SE and the input image. The size and shape of the SE play a crucial role in determining the outcome of image processing. Given an original image $f$ and an SE ${b}$, the basic morphological dilation operation $\delta_{{b}}(f)=f\oplus{b}$ and erosion operation $\varepsilon_{{b}}(f)=f\ominus{b}$ are defined as follows:
\begin{equation}\begin{cases}
\delta_{{b}}(f)=\max_{(s,t)\in D_{{b}}}\{f(x-s,y-t)\} \\
\varepsilon_{{b}}(f)=\min_{(s,t)\in D_{{b}}}\{f(x+s,y+t)\} 
\end{cases},\end{equation}
where $D_{{b}}$ is the domain of the structuring element ${b}$. In the dilation operation, the maximum pixel value within the region covered by the structuring element is taken as the new pixel value for that position. Conversely, in the erosion operation, the minimum pixel value within the covered region is used.

\textit{2) Grayscale Morphological Reconstruction (GMR)}:  
Morphological reconstruction uses two images: the marker image and the mask image, along with a structuring element. It performs a series of morphological operations to modify the marker image until it satisfies certain conditions of the mask image. The basic operations of morphological reconstruction include reconstruction based on geodesic dilation and geodesic erosion.

Let \( f \) be the marker image, \( g \) be the mask image, and \( b \) be the structuring element. The \( k \)-th order geodesic dilation \( \delta_{g}^{(k)}(f) \) $(k \geq1)$ is defined as:
\begin{equation}\begin{aligned}
 & \delta_{g}^{(k)}(f) = \delta_b\left( \delta_{g}^{(k-1)}(f) \right) \wedge g \quad (k \geq 2), \\
 & \delta_{g}^{(1)}(f) = \delta_b(f) \wedge g,
\end{aligned}\end{equation}
where the $\wedge$ operator denotes pointwise minimum. 

Consistent with the definitions provided above, the \( k \)-th order geodesic erosion \( \varepsilon_{g}^{(k)}(f) \)$(k \geq1)$ is expressed as:
\begin{equation}\begin{aligned}
 & \varepsilon_{g}^{(k)}(f) = \varepsilon_b\left( \varepsilon_{g}^{(k-1)}(f) \right) \vee g \quad (k \geq 2), \\
 & \varepsilon_{g}^{(1)}(f) = \varepsilon_b(f) \vee g,
\end{aligned}\end{equation}
where $\vee$ represents pointwise maximum.

The reconstruction \( R_{g}(f) \) of \( f \) with respect to \( g \) is achieved by repeatedly applying geodesic dilation or geodesic erosion operations until the image no longer changes, i.e.,
\begin{equation}\begin{aligned}
 & R_{g}^{\delta}(f) = \lim_{k \to \infty} \delta_{g}^{(k)}(f). \\
 & R_{g}^{\varepsilon}(f) = \lim_{k \to \infty} \varepsilon_{g}^{(k)}(f).
\end{aligned}\end{equation}
The overall flowchart of GMR is shown in the lower right corner of Fig. 1. GMR begins by performing an erosion operation on the infrared image. The original image \( t \) serves as the mask image, and the result of the erosion operation \( \varepsilon_{{b}}(t) \) is used as the marker image. The image \( t \) is then reconstructed by iteratively applying unit geodesic dilation to \( \varepsilon_{{b}}(t) \).
The erosion operation \( \varepsilon_{{b}}(t) \) reduces high-intensity areas and increases low-intensity areas. This operation helps remove negative connected components in the image during reconstruction.

To further eliminate interference from positive connected components, a dilation operation is applied to the reconstructed image \( S = R_{t}^{\delta}[\varepsilon_{{b}}(t)] \), denoted as \( \delta_{b}(S) \). The image \( S \) serves as the mask, and the result of the dilation operation \( \delta_{b}(S) \) is used as the marker image. The image \( \delta_{b}(S) \) is then reconstructed by applying the unit geodesic erosion operation to \( S \) iteratively until stability is reached.The dilation operation reduces low-intensity areas and increases high-intensity areas, effectively eliminating both negative and positive connected components in the image during reconstruction. The final result of GMR is denoted as \( R_{S}^{\varepsilon}[\delta_{b}(S)] \).

\textit{3) Difference of GMR}:
In morphological processing theory, the structuring element (SE) can act as a probe for extracting regions of interest from an image, and its construction directly influences the resulting marker image. Since noise typically consists of isolated pixels with significant gray-level differences from their neighboring pixels, small structuring elements (e.g., 2 × 2 in size) are used to eliminate both positive and negative noise in the original image. Subsequently, GMR with larger structuring elements can be applied to remove large bright and dark connected regions in the original image.

Thus, we subtract the GMR result using a 2 × 2 structural element from the GMR result using an $N \times N$ large-scale structural element to obtain the differential GMR result. Next, we divide it into two parts with 0 as the boundary. The part greater than 0 is extracted to construct the bright region saliency map, and the part smaller than 0 is extracted to construct the dark region saliency map, effectively extracting multi-scale saliency information from the image. Finally, the extracted bright saliency map $f_{m}^{b}$ and dark saliency map $f_{m}^{d}$ are linearly fused with the base image $f_{B}$ obtained from the GMR with a 2 × 2 structural element to generate the final enhanced infrared image $f_{E}$, as shown below:
\begin{equation}f_{E}=f_{B}+\alpha_{1}\times\frac{f_{m}^{d}}{\max\left(\left|f_{m}^{d}\right|\right)}+\alpha_{2}\times\frac{f_{m}^{b}}{\max\left(f_{m}^{b}\right)},\end{equation}
where $\alpha_{1}$ and $\alpha_{2}$ are the fusion coefficients for the dark and bright saliency maps, respectively.

\begin{figure*}[h!]
    \centering
    \vspace{0.3cm}
    \includegraphics[width=0.9\textwidth]{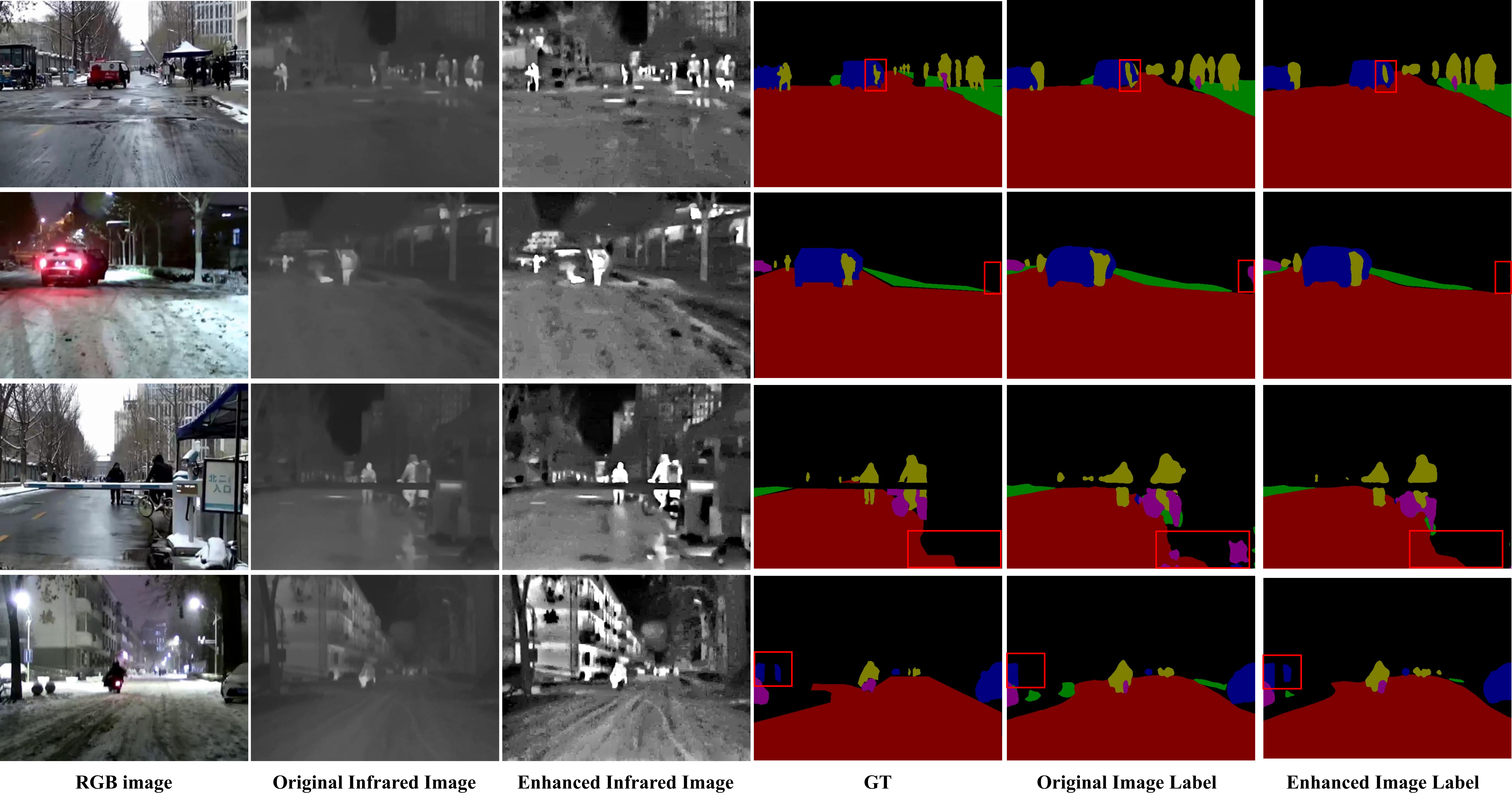}
    \caption{Visualization results of semantic segmentation of our proposed method on SUS dataset, where the red box indicates the superiority of our method}
    \label{figurelabel}
\end{figure*}

   \begin{figure}[thpb]
    \centering
    \vspace{0.3cm}
    
        \includegraphics[width=0.8\columnwidth]{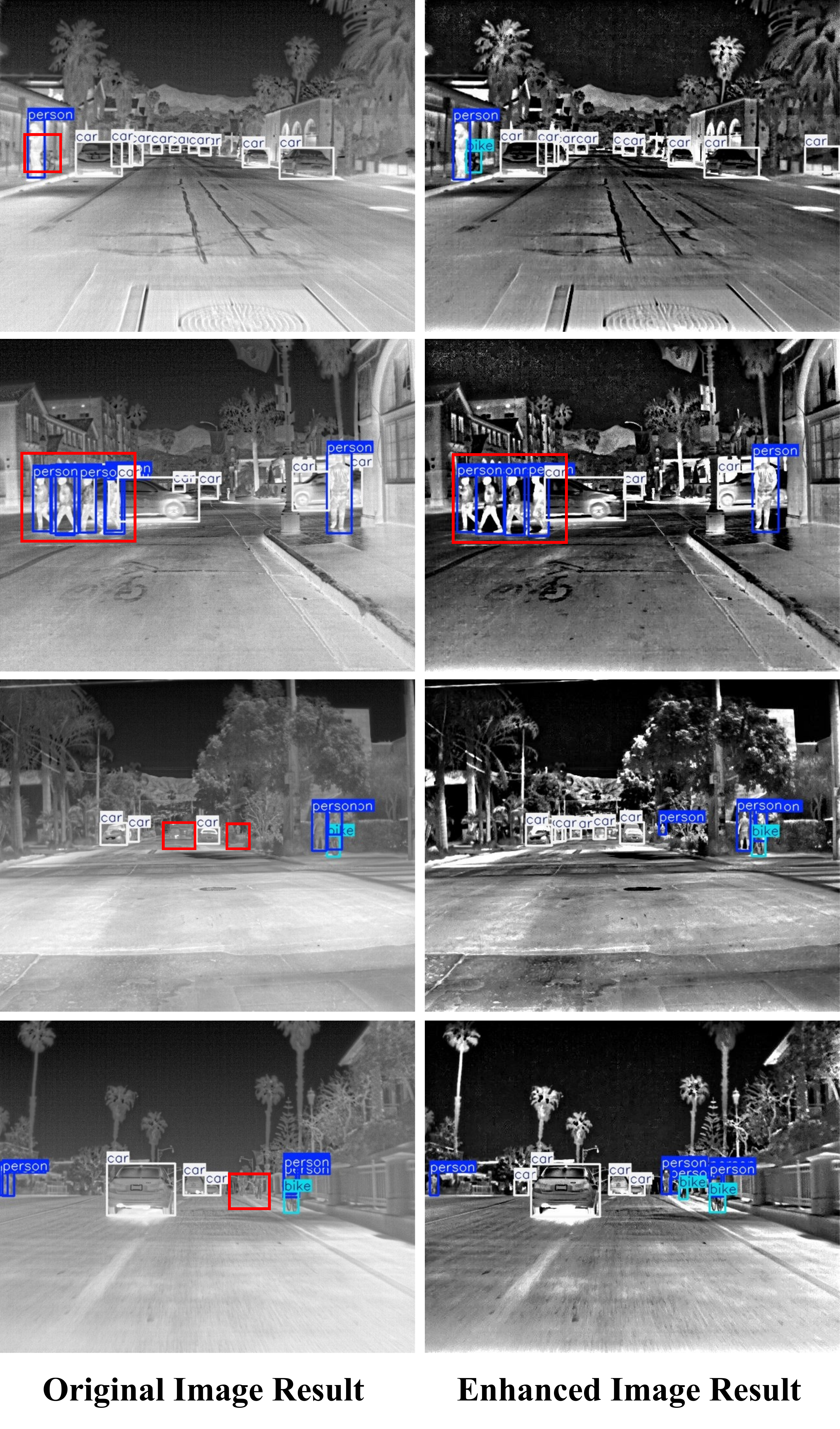}
    \caption{Visualization results of our proposed method for target detection on FLIR dataset, where the red box indicates the superiority of our method}
    \label{figurelabel}
\end{figure}

\section{Experiment}
\subsection{Experimental Settings}

{\bf Datasets}. We evaluate the proposed algorithm using two publicly available urban road datasets that include extreme conditions such as low light and snow: the FLIR object detection dataset and the SUS semantic segmentation dataset, to assess the preprocessing effects of our enhancer on high-level vision tasks.

1) FLIR Dataset: The FLIR dataset is a multispectral road scene dataset containing infrared and RGB images. A ``clean version" of the FLIR dataset was later introduced, removing incorrectly annotated thermal images \cite{zhang2020multispectral}. In this work, we use the ``clean version" of the front-view infrared dataset, which contains 4,129 training images and 1,013 test images. It includes three annotated categories: ``person", ``bicycle", and ``car".

2) SUS Dataset\cite{guo2025transferring}: The SUS dataset is an RGB-T road scene dataset captured under snowy conditions. It contains 1,035 pairs of aligned RGB-T images captured in snowy urban scenes, including 503 pairs of daytime images and 532 pairs of nighttime images. The SUS dataset is divided into training, validation, and test sets, with proportions of 70\%, 10\%, and 20\%, respectively. It includes five annotation labels: ``road", ``sidewalk", ``person", ``vehicle", and ``bicycle".

{\bf Metrics}. For the object detection task, we perform quantitative evaluation using standard metrics, including mean Average Precision (mAP) and mAP$_{50}$. mAP is calculated as the average over all categories with an IoU threshold ranging from 0.50 to 0.95 at intervals of 0.05. For the semantic segmentation task, we employ mean intersection over union (mIoU) as the evaluation metric to assess segmentation performance.

\textbf{Implementation details}. For the proposed algorithm, in the layer decomposition process, the $l_0$-$l_1$ model parameters are $\lambda_1 = 0.3$, $\lambda_2 =\lambda_1\times0.01$, and in the linear fusion process, $\alpha = 1.2$, $\beta = 0.8$. In the morphological reconstruction process, using too large a structuring element results in a decrease in accuracy, while too small a structuring element increases the computational load. Therefore, the large-scale structuring element is set as $N = \min \{ h, w \} / 5$, where $h$ and $w$ represent the height and width of the image, respectively. Then, in the linear fusion part, $\alpha_1$ is the minimum value of the base layer image, and $\alpha_2$ is 255 minus the maximum value of the base layer image.

We conduct experiments on high-level vision tasks using the PyTorch library for training, evaluation, and testing. All experiments are performed on an NVIDIA 4090 GPU. We retrain downstream task models using the enhanced images.
For the object detection task, we employ the YOLOv11 model \cite{Jocher_Ultralytics_YOLO_2023} for testing, utilizing the YOLOv11n model structure without initial pretrained weights. Training is optimized using SGD with a batch size of 32 and num\_workers of 8.  
For the semantic segmentation task, we utilize the CLNet-T model \cite{guo2024contrastive}. Before training, the SegFormer-B4 backbone is initialized with pretrained weights from the ImageNet dataset, whereas the remaining parts are randomly initialized. Random cropping and flipping are used to augment the data during training. Training is optimized using Ranger with a weight decay of \(5 \times 10^{-4}\). For CLNet-T, we set the batch size to 4, the num\_workers to 4, and the initial learning rate to \(1 \times 10^{-4}\).

\begin{table}[ht]
\caption{Quantitative results of various methods on the FLIR object detection dataset, with bold indicating the best results.}
\label{table_example}
\centering
\resizebox{\columnwidth}{!}{%
\begin{tabular}{lcccccccc}
\toprule
\multirow{2}{*}{\textbf{method}} & \multicolumn{2}{c}{\textbf{person}} & \multicolumn{2}{c}{\textbf{bike}} & \multicolumn{2}{c}{\textbf{car}} & \multicolumn{2}{c}{\textbf{all}} \\ \cmidrule(lr){2-3} \cmidrule(lr){4-5} \cmidrule(lr){6-7} \cmidrule(lr){8-9}
                                & \textbf{mAP$_{50}$} & \textbf{mAP}  & \textbf{mAP$_{50}$} & \textbf{mAP}  & \textbf{mAP$_{50}$} & \textbf{mAP}  & \textbf{mAP$_{50}$} & \textbf{mAP}  \\ \midrule
original                       & 0.767         & \textbf{0.372}         & 0.453          & 0.162          & \textbf{0.881}          & \textbf{0.583}          & 0.7          & 0.372          \\
CLAHE                          & 0.717         & 0.33          & 0.402          & 0.138          & 0.873          & 0.566          & 0.664         & 0.345         \\
IE-CGAN                       & 0.738         & 0.341         & 0.484          & 0.166          & 0.874          & 0.575          & 0.699         & 0.361         \\
fieldscale                     & \textbf{0.769 }        & 0.365         & 0.453          & 0.155          & 0.879          & 0.576          & 0.7          & 0.365         \\
IGIM                           & 0.765         & 0.36          & 0.491          & 0.171          & 0.876          & 0.574          & 0.711         & 0.369         \\
ours                           & 0.764 & 0.367 & \textbf{0.502}  & \textbf{0.186}  & 0.874  & 0.575  & \textbf{0.714}  & \textbf{0.376}  \\ \bottomrule
\end{tabular}
}
\end{table}

\begin{table}[thpb]
\caption{Test results for all methods on the validation set of the SUS dataset. Where bold indicates the best performance}
\label{table_example}
\centering
\resizebox{\columnwidth}{!}{%
\begin{tabular}{lcccccc}
\toprule
\multirow{2}{*}{\textbf{method}} & \multicolumn{5}{c}{\textbf{IoU}} & \multirow{2}{*}{\textbf{mIoU}} \\ \cmidrule(lr){2-6}
                                & \textbf{road} & \textbf{sidewalk} & \textbf{person} & \textbf{vehicle} & \textbf{bicycle} & \\ \midrule
original                       & 0.949         & \textbf{0.702}             & 0.744           & 0.882            & 0.767           & 0.809         \\
CLAHE                          & 0.945         & 0.652             & 0.717           & 0.880            & 0.779           & 0.795         \\
IE-CGAN                       & \textbf{0.950}         & \textbf{0.702}             & 0.756           & 0.884            & 0.765           & 0.811         \\
fieldscale                     & \textbf{0.950}         & 0.684             & \textbf{0.772}           & 0.874            & 0.770           & 0.810         \\
IGIM                           & 0.948         & 0.661             & 0.757           & 0.880            & 0.787           & 0.807         \\
ours                           & 0.949 & 0.663     & 0.767   & \textbf{0.886}    & \textbf{0.798}  & \textbf{0.813} \\ \bottomrule
\end{tabular}
}
\end{table}

\begin{table}[thpb]
\caption{Test results for all methods on the test set of the SUS dataset. Where bold indicates the best performance}
\label{table_example}
\centering
\resizebox{\columnwidth}{!}{%
\begin{tabular}{lcccccc}
\toprule
\multirow{2}{*}{\textbf{method}} & \multicolumn{5}{c}{\textbf{IoU}} & \multirow{2}{*}{\textbf{mIoU}} \\ \cmidrule(lr){2-6}
                                & \textbf{road} & \textbf{sidewalk} & \textbf{person} & \textbf{vehicle} & \textbf{bicycle} & \\ \midrule
original                       & 0.949         & 0.677             & 0.756           & 0.883            & 0.710           & 0.795         \\
CLAHE                          & 0.945         & 0.676             & 0.733           & 0.868            & 0.687           & 0.782         \\
IE-CGAN                       & 0.947         & 0.675             & 0.769           & \textbf{0.887}            & 0.714           & 0.798         \\
fieldscale                     & \textbf{0.951}         & 0.690             & \textbf{0.781}           & 0.882            & 0.712           & 0.803         \\
IGIM                           & 0.948         & 0.667             & 0.763           & 0.882            & 0.697           & 0.791         \\
ours                           & 0.948 & \textbf{0.692}     & 0.772   & 0.886    & \textbf{0.720}  & \textbf{0.804} \\ \bottomrule
\end{tabular}
}
\end{table}

\begin{table}[thpb]
\caption{Ablation experiment of our proposed method for target detection on FLIR dataset}
\label{table_example}
\centering
\resizebox{\columnwidth}{!}{%
\begin{tabular}{cccc}
\toprule
\textbf{Layer Decomposition} & \textbf{Salient Information Extraction} & \textbf{mAP$_{50}$} & \textbf{mAP} \\
\midrule
\checkmark & & 0.711 & 0.371 \\
 & \checkmark & 0.703 & 0.373 \\
\checkmark & \checkmark & \textbf{0.714} & \textbf{0.376} \\
\bottomrule
\end{tabular}
}
\end{table}

\begin{table}[htbp]
  \centering
  \caption{Ablation experiment of our proposed method on SUS dataset}
  \label{table_example}
  \begin{tabular}{l S[table-format=2.3] S[table-format=2.3] S[table-format=2.3]}
    \toprule
    {Component / Metric} & {Config1} & {Config2} & {Config3} \\
    \midrule
    Layer Decomposition      & $\checkmark$ &   & $\checkmark$ \\
    Salient Information Extraction &   & $\checkmark$ & $\checkmark$ \\
    \midrule
    EN ↑ & \text{6.936} & \text{6.299} & \textbf{7.312} \\
    SF ↑ & \text{7.129} & \text{5.954} & \textbf{12.434} \\
    AG ↑ & \text{1.998} & \text{1.778} & \textbf{4.005} \\
    SD ↑ & \text{42.926} & \text{30.756} & \textbf{54.458} \\
    VIF ↑ & \text{3.488} & \text{2.812} & \textbf{3.626} \\
    \bottomrule
  \end{tabular}
\end{table}

\subsection{Comparisons with SOTA Methods}

\textbf{Quantitative Evaluation.} We conduct a comprehensive quantitative comparison between the proposed method and SOTA methods, including CLAHE \cite{mohan2012modified}, IE-CGAN \cite{kuang2019single}, FieldScale \cite{gil2024fieldscale}, and IGIM \cite{wang2024raw}, across object detection and semantic segmentation tasks.

Table I summarizes the quantitative evaluation results for the object detection task. On the FLIR\_align dataset, the proposed method demonstrates superior performance compared to existing infrared image enhancement algorithms, achieving significant improvements of 1.4\% in mAP and 0.6\% in mAP$_{50}$ over the original images. Notably, our method achieves a remarkable enhancement in the bicycle category, with a 2.4\% increase in mAP and a 4.9\% increase in mAP$_{50}$.
Tables II and III summarize the quantitative evaluation results for the semantic segmentation task. On the SUS dataset, our method consistently outperforms other infrared image approaches, achieving mIoU improvements of 0.4\% and 0.9\% on the validation and test sets, respectively, compared to the original images. Meanwhile, our method achieves improvements in metrics across all target categories relevant to downstream tasks. Particularly significant are the notable accuracy improvements in the vehicle and bicycle categories. The experimental results from these advanced vision tasks demonstrate that our method enhances the contrast of targets required for downstream tasks, particularly for non-self-heating objects such as bicycles. The generated infrared images are more suitable for high-level vision tasks, and these comprehensive experimental results confirm the effectiveness of the proposed algorithm.

In addition, through the aforementioned quantitative experiments, we also found that due to the significant changes in the properties of the infrared images after enhancement, the enhanced algorithm may have better visual effects, but it does not necessarily improve the metrics of downstream tasks. Sometimes, the change in properties even leads to worse accuracy than the original images. We believe this is due to the inherent task differences between the two tasks. At the same time, our algorithm has minimized this gap as much as possible.

\textbf{Qualitative Evaluation}. 
Fig. 3 compares the infrared image enhancement results of our method with other approaches on the FLIR\_align and SUS datasets. As shown, CLAHE and IGIM achieve notable contrast improvement but introduce significant noise. IE-CGAN exhibits issues of overexposure and underexposure, whereas Fieldscale suffers from the loss of features in dark regions. In contrast to other infrared image enhancement algorithms, our method preserves more details and highlights targets more prominently without introducing noise.

Fig. 4 illustrates the semantic segmentation results on the SUS dataset. Our method reduces segmentation errors for bicycles and improves segmentation accuracy for pedestrians and bicycles, resulting in outputs that are closer to the ground truth. Fig. 5 showcases the object detection results on the FLIR\_align dataset using our proposed method. The enhanced images reduce overlapping bounding boxes in pedestrian detection and identify pedestrians and bicycles that were missed in the original images. These qualitative results demonstrate the superiority of the infrared images generated by our method for high-level vision tasks, thereby validating the effectiveness of our approach.

\subsection{Ablation Studies and Analysis}

To evaluate the effectiveness of each component in the proposed infrared image enhancement framework, we conducted ablation studies on the FLIR\_align dataset. Since our infrared image enhancement algorithm is designed to meet the requirements of high-level vision tasks, we used object detection accuracy as the metric to evaluate the effectiveness of each module. We applied the layer decomposition and saliency information extraction components independently for object detection. Table IV summarizes the results of our ablation experiments. Compared to the original images, object detection accuracy improved when using only layer decomposition or saliency information extraction, but it remained lower than the accuracy achieved by our full algorithm.

Additionally, we employed image evaluation metrics to further validate the effectiveness of each component. Specifically, we utilized metrics including information entropy (EN), spatial frequency (SF), average gradient (AG), standard deviation (SD), and visual information fidelity (VIF). Table V presents the results of the ablation experiments. Our algorithm outperformed the single-module versions across these metrics, further validating the superiority of our infrared image enhancement framework.

\section{CONCLUSIONS}

In this paper, we proposed a task-oriented infrared image enhancement method that highlighted key targets, including pedestrians, vehicles, and bicycles, leading to improvements in the metrics of high-level vision tasks such as object detection and semantic segmentation. Our approach comprised two steps: layer decomposition and saliency information extraction. First, we applied an $l_0$-$l_1$ layer decomposition technique for infrared images, which enhanced scene details while preserving important information, such as features in dark regions, thereby providing more information for the subsequent saliency extraction step. Next, using a morphological reconstruction-based saliency extraction method, we effectively extracted and enhanced target information from the scene without amplifying noise, thereby improving perception for high-level vision tasks. Extensive experiments demonstrated that our method outperformed SOTA methods, highlighting its potential to enhance the performance of autonomous driving systems in complex environments.

\bibliographystyle{IEEEtran}
\bibliography{IEEEexample}

\end{document}